\documentclass{article}

\usepackage{wrapfig}
\usepackage{booktabs}
\usepackage{microtype}
\usepackage{graphicx}
\usepackage{subcaption}
\usepackage{booktabs} % for professional tables
\usepackage{hyperref}
\usepackage{tabularray}
\UseTblrLibrary{booktabs}
\usepackage{xcolor}
\usepackage{colortbl}
\usepackage{pifont}  % for \ding command
\usepackage{enumitem}
\usepackage{tcolorbox}
\usepackage{ulem}
\usepackage{xspace}
\usepackage{bbold}
\usepackage{url}
\definecolor{LightCyan}{rgb}{0.8, 0.9, 1}
\definecolor{maroon}{cmyk}{0,0.87,0.68,0.32}
\newcolumntype{a}{>{\columncolor{LightCyan}}c}

\newcommand{\method}{\texttt{Med-TIV}\xspace}
\usepackage[preprint]{icml2026}

\usepackage{amsmath}
\usepackage{amssymb}
\usepackage{mathtools}
\usepackage{amsthm}
\usepackage{pifont}
\usepackage{fontawesome5}
\usepackage{listings}
\newcommand{\cmark}{\ding{51}}  % ✓
\newcommand{\xmark}{\ding{55}}  % ✗
\newcommand{\tool}{\faWrench}   % ��

\usepackage[capitalize,noabbrev]{cleveref}

\theoremstyle{plain}

\theoremstyle{definition}

\theoremstyle{remark}

\usepackage[textsize=tiny]{todonotes}

\icmltitlerunning{Scaling Medical Reasoning Verification via Tool-Integrated Reinforcement Learning}

\begin{document}

\twocolumn[
  \icmltitle{Scaling Medical Reasoning Verification via Tool-Integrated Reinforcement Learning}



  \icmlsetsymbol{equal}{*}

  \begin{icmlauthorlist}
    \icmlauthor{Hang Zhang}{pitt}
    \icmlauthor{Ruheng Wang}{utsw}
    \icmlauthor{Yuelyu Ji}{pitt}
    \icmlauthor{Mingu Kwak}{pitt}
    \icmlauthor{Xizhi Wu}{pitt}
    \icmlauthor{Chenyu Li}{pitt}
    \icmlauthor{Li Zhang}{pitt}
    \icmlauthor{Wenqi Shi}{utsw}
    \icmlauthor{Yifan Peng}{cornell}
    \icmlauthor{Yanshan Wang}{pitt}
  \end{icmlauthorlist}

  \icmlaffiliation{pitt}{University of Pittsburgh}
  \icmlaffiliation{utsw}{UT Southwestern Medical Center}
  \icmlaffiliation{cornell}{Cornell University}

  \icmlcorrespondingauthor{Hang Zhang}{haz269@pitt.edu}
  \icmlcorrespondingauthor{Yanshan Wang}{yaw45@pitt.edu}

  \icmlkeywords{Machine Learning, ICML}

  \vskip 0.3in
]



\printAffiliationsAndNotice{}  % no special notice (required even if empty)

\begin{abstract}
   Large language models have achieved strong performance on medical reasoning benchmarks, yet their deployment in clinical settings demands rigorous verification to ensure factual accuracy. While reward models offer a scalable approach for reasoning trace verification, existing methods face two limitations: they produce only scalar reward values without explicit justification, and they rely on single-pass retrieval that precludes adaptive knowledge access as verification unfolds. We introduce $\method$, an agentic framework that addresses these limitations by training medical reasoning verifiers to iteratively query external medical corpora during evaluation. Our approach combines tool-augmented verification with an iterative reinforcement learning paradigm that requires only trace-level supervision, alongside an adaptive curriculum mechanism that dynamically adjusts training data distribution. Across four medical reasoning benchmarks, $\method$ achieves substantial gains over existing methods, improving MedQA accuracy by 23.5\% and MedXpertQA by 32.0\% relative to the base generator in particular. Crucially, $\method$ demonstrates an $\mathbf{8\times}$ reduction in sampling budget requirement compared to prior reward model baselines. These findings establish that grounding verification in dynamically retrieved evidence offers a principled path toward more reliable medical reasoning systems.
\end{abstract}

\section{Introduction}

\begin{figure}[t]
    \centering
    \includegraphics[width=\columnwidth]{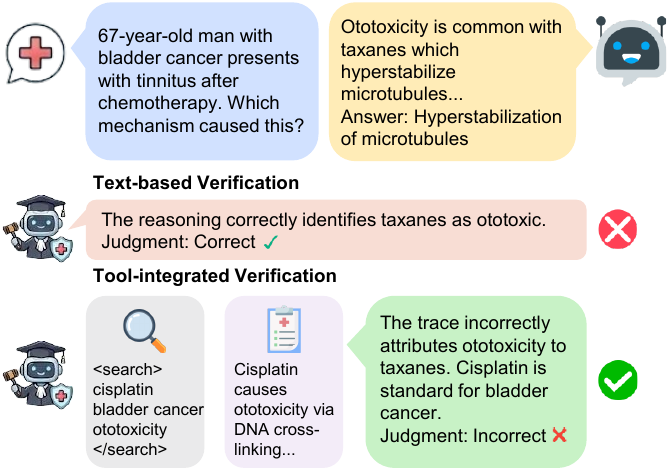}
    \caption{Comparison of medical reasoning verification paradigms. Text-based judges rely on parametric knowledge and may validate erroneous reasoning, while tool-integrated judges dynamically retrieve evidence to ground their judgments.}
    \label{fig:example}
\end{figure}

Large Language Models (LLMs) have demonstrated remarkable capabilities in medical reasoning, achieving competitive performance on clinical question answering, diagnostic inference, and medical knowledge benchmarks~\cite{ji2025mitigating, XIAO2026109184}. While these advances hold significant promise for augmenting clinical decision making and democratizing access to medical expertise, the deployment of LLMs in high-stakes clinical settings demands rigorous verification mechanisms to ensure that generated reasoning is both factually accurate and logically sound~\cite{zhang2025safeaiclinicianscomprehensive, wang2025digital}.

Reward-based judges have therefore emerged as a scalable solution for evaluating model outputs, supporting both post-training refinement via reinforcement learning from human feedback (RLHF) and inference-time scaling through tree search~\cite{snell2024scalingllmtesttimecompute}. These judges can be broadly categorized by the granularity of their supervision. Outcome Reward Models (ORMs) provide sparse trace-level supervision that quantifies the quality of the entire output, while Process Reward Models (PRMs) offer dense step-level feedback that scores each intermediate reasoning step, enabling fine-grained credit assignment and precise error localization within multi-step reasoning. Recent work has adapted both paradigms to the medical domain to assess complex clinical reasoning traces. In parallel, advances in generative reward modeling have extended judge models beyond scalar scoring, enabling them to produce natural-language critiques that explicitly justify their decisions~\cite{liu2025inferencetimescalinggeneralistreward, xiong2025stepwiserstepwisegenerativejudges}.

Despite their effectiveness, reward-based judges exhibit fundamental limitations when applied to clinical reasoning tasks~\cite{yun2025medprmmedicalreasoningmodels}. 
A primary concern is the prevalence of hallucinations in critique traces, where judge models generate plausible yet factually incorrect assessments (Figure \ref{fig:example}). This issue is particularly noticeable in the medical domain, where reliable verification demands grounding in authoritative clinical evidence and established medical knowledge. Unverified judgments could lead to the propagation of incorrect diagnostic or treatment recommendations. 
%
% Existing work on medical reasoning verifiers, such as Med-PRM~\cite{yun2025medprmmedicalreasoningmodels}, suffers from several notable limitations. 
Existing medical reasoning verifiers typically provide only scalar reward signals, offering little or no justification for their judgments and thus limiting interpretability~\cite{jiang2025meds3medicalslowthinking}. Furthermore, these methods often rely on a static Retrieval-Augmented Generation (RAG) pipeline, in which a fixed set of retrieved documents is prefixed to the context and remains unchanged throughout evaluation~\cite{yun2025medprmmedicalreasoningmodels}. Such static design precludes adaptive, multi-turn evidence gathering and forces the verifier to a fixed retrieval budget, thus limiting scalability. 

To address these issues, we propose \method (\textbf{Med}ical \textbf{T}ool-\textbf{I}ntegrated reasoning \textbf{V}erifier), an agentic reinforcement learning (RL) framework that trains LLMs to leverage external knowledge bases for judging medical reasoning traces\footnote{Code is available at \url{https://github.com/PittNAIL/med-tiv}}. \method features three key design principles: (1) a tool-augmented verification paradigm that enables dynamic, iterative knowledge retrieval during the evaluation process; (2) an iterative RL approach that progressively improves verification capabilities without requiring step-level expert annotations; and (3) an adaptive curriculum formulation strategy that adjusts the data distribution in response to the evolving capability of the model. By equipping judge models with tool-use capabilities, $\method$ grounds evaluation decisions in external evidence rather than relying solely on parametric knowledge, thereby mitigating hallucination, improving interpretability, and overcoming the limitations of static RAG~\cite{ji2025biasevaluationmitigationretrievalaugmented, xia2025parallelismmeetsadaptivenessscalable}.

To verify the effectiveness of \method, we conduct extensive experiments on common medical reasoning benchmarks. Our results demonstrate that \method trains strong medical verifiers: when guiding inference-time search for a 7B generator model, our trained verifier achieves relative improvements of 23.5\% on MedQA and 32.0\% on MedXpertQA compared to the generator model alone. Moreover, \method consistently outperforms existing medical reward model baselines and surpasses the performance of models that are up to $\mathbf{4\times}$ larger in scale. Notably, \method also demonstrates an $\mathbf{8\times}$ gain in sampling efficiency compared to prior reward-based approaches, achieving equivalent accuracy with substantially fewer sampled reasoning traces during test-time search.

Our main contributions are summarized as follows:
\begin{itemize}[leftmargin=*,nosep]
    \item We propose \method, a novel tool-integrated verification framework that enables dynamic, iterative knowledge retrieval during medical reasoning evaluation, providing both interpretable, fine-grained justification and improved factual grounding.
    \item We introduce an iterative RL paradigm with curriculum-based difficulty adaptation that progressively improves verification capabilities through self-bootstrapping, requiring only trace-level supervision rather than dense step-level expert annotations.
    \item \method achieves state-of-the-art performance on four medical reasoning benchmarks, with comprehensive ablation studies that validate each component's contribution.
\end{itemize}

\section{Preliminaries}

\subsection{Problem Setup}

We define \textit{medical reasoning verification} as the task of assessing the correctness of a multi-step reasoning trace generated in response to a medical question. Formally, given a medical question $q \in \mathcal{Q}$ and a multi-step reasoning trace $\tau = (s_1, s_2, \ldots, s_m)$ from a generator model, a verifier model determines whether $\tau$ contains any errors. We formulate this problem as binary classification, where the verifier $V_\theta(q, \tau)$ produces a judgment
$
\ell \in \{0, 1\},
$
where $\ell = 1$ indicates a error-free reasoning trace, and $\ell = 0$ indicates the presence of one or more errors. Unlike scalar reward models that output continuous scores, we adopt a generative judge paradigm in which the verifier produces a discrete judgment accompanied by a detailed critique trace that provides a structured justification for the decision.

\begin{figure*}[t]
    \centering
    \begin{subfigure}[b]{0.95\textwidth}
        \centering
        \includegraphics[width=\linewidth]{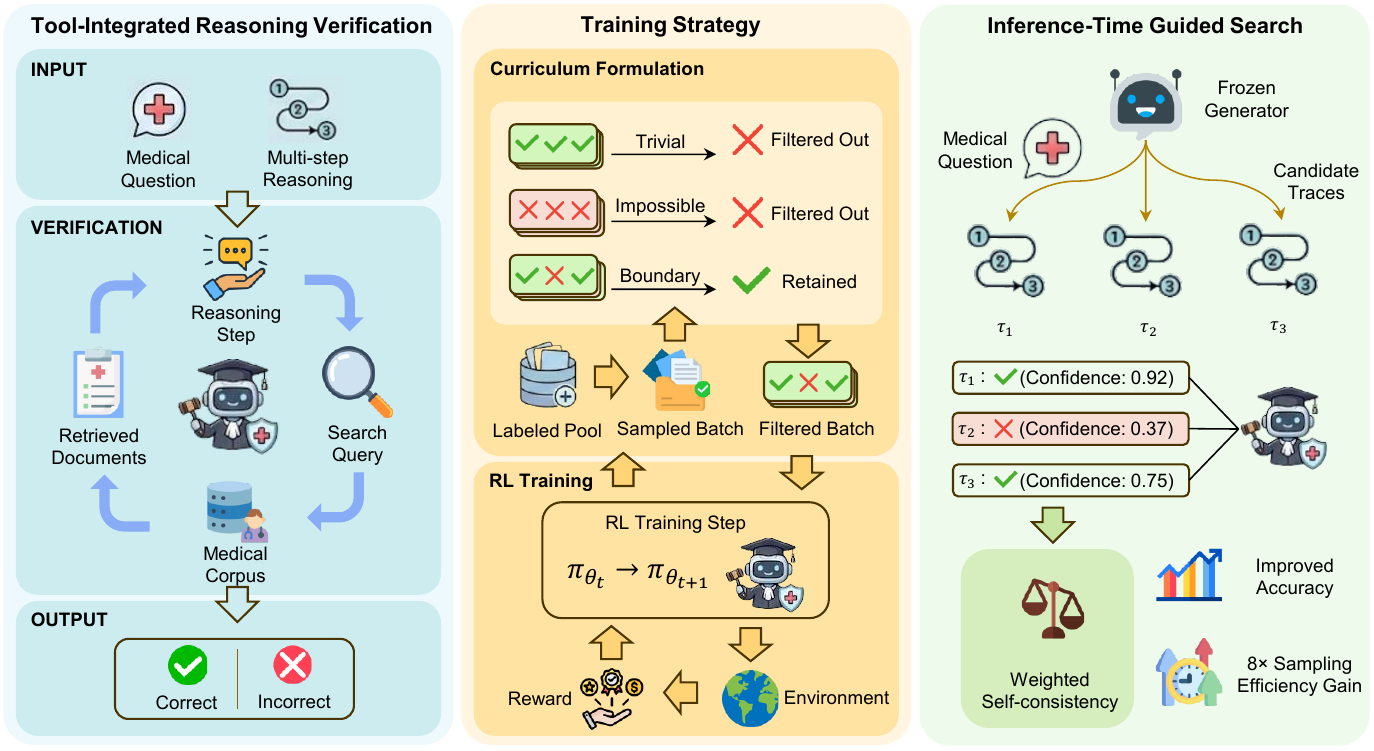}
    \end{subfigure}
    \caption{\textbf{Overview of \method.} \textit{Left:} Tool-integrated verification iteratively analyzes reasoning traces, formulates search queries, and retrieves medical evidence before producing correctness judgments. \textit{Middle:} Curriculum formulation filters trivial and impossible instances, retaining boundary cases for RL training. \textit{Right:} At inference time, the verifier evaluates candidate medical reasoning traces generated by a frozen model and final answers are selected via weighted self-consistency.}
    \label{fig:framework}
\end{figure*}

\subsection{Tool-Augmented Reasoning Verifier}

Following prior works~\cite{jin2025searchr1trainingllmsreason}, we extend the verifier with access to an external \textit{search engine} $\mathcal{E}$ that retrieves top-$k$ documents from a curated medical corpus (See Appendix~\ref{sec:retrieval_setup} for details). Retrieved documents are appended verbatim to the verifier context. Given a verification instance $(q, \tau)$, the verifier constructs an iterative verfication trajectory. At step $k$, the trajectory is represented as:
\[
    \mathbf{t}_k = \{r_1, a_1, o_1, \ldots, r_k, a_k, o_k\},
\]
where $r_i$ denotes a natural language reasoning step analyzing the medical content, $a_i$ is a search query formulated to retrieve relevant medical knowledge, and $o_i = \mathcal{E}(a_i)$ represents the retrieved documents. The iterative verification process is defined as:
\[
    (r_k, a_k) \sim V_\theta(q, \tau, \mathbf{t}_{k-1}),
\]
\[
    o_k = \mathcal{E}(a_k),
\]
\[
    \mathbf{t}_k = \mathbf{t}_{k-1} \oplus r_k \oplus a_k \oplus o_k,
\]
where $\oplus$ denotes sequence concatenation. This process continues until the verifier produces a final judgment $\ell \sim V_\theta(q, \tau, \mathbf{t}_T)$ at the terminal step $T$. By allowing multiple tool executions, the verifier dynamically retrievs medical knowledge as need to verify specific claims in the reasoning trace. Table~\ref{tab:prompt} in the Appendix shows the explicit instruction used in our experiments.

\subsection{Test-Time Search}

Test-time search strategies improve reasoning performance by leveraging reward models to evaluate and select among multiple candidate solutions~\cite{shi2024medadapterefficienttesttimeadaptation}. Given a frozen generator model $\pi_{\text{gen}}$ and a question $q$, we first sample $N$ independent reasoning traces:
\[
    \{\tau^{(j)}\}_{j=1}^N \sim \pi_{\text{gen}}(\cdot \mid q).
\]
A trained verifier $V_\theta$ then scores each candidate trace, and the final output is selected based on these scores. Common selection strategies include Best-of-$N$ sampling, which selects the trace with the highest score:
\[
    \hat{\tau} = \arg\max_{\tau^{(j)}} V_\theta(q, \tau^{(j)}),
\]
and verification-based majority voting, where candidate traces are first filtered by the verifier and the final answer is determined by consensus among verified traces. \method trains such a plug-in verifier that provides tool-grounded assessments that can be used to augment decision-making for any frozen generator model at inference time.

\section{Tool-Integrated Medical Reasoning Verifier}

\method is an agentic verification framework that trains models to leverage external knowledge bases for verifying whether a given medical reasoning trace contains errors. We adopt an iterative training approach based on dynamic curriculum learning, which requires no fine-grained step-level expert supervision and trains solely through multiple rounds of reinforcement learning (Figure \ref{fig:framework}). We next describe the training procedure of \method in details.

\subsection{Tool-Integrated RL with Verifiable Rewards}

\paragraph{Data Construction.}

All training data across iterations is derived from the open-source Med-PRM dataset. Each original instance consists of a tuple $(q, \tau, \ell_{\text{step}}, \ell_{\text{trace}})$, where $q$ is a medical question, $\tau$ is a multi-step reasoning trace, $\ell_{\text{step}}$ denotes step-level labels, and $\ell_{\text{trace}}$ is a trace-level correctness label\footnote{Dataset is available at \url{https://huggingface.co/datasets/dmis-lab/llama-3.1-medprm-reward-training-set}}.

At each training iteration, we only utilize the triplet $(q, \tau, \ell_{\text{trace}})$ with human-annotated trace-level labels. Step-level labels $\ell_{\text{step}}$ is intentionally excluded, as \method is designed to improve verification performance without replying on fine-grained supervision.
For each training iteration, we fix the training data budget to 20K instances and enforce a balanced label distribution between correct ($\ell_{\text{trace}} = 1$) and incorrect ($\ell_{\text{trace}} = 0$) reasoning traces.

\paragraph{Algorithm.}

We employ Dr. GRPO ~\cite{liu2025understandingr1zeroliketrainingcritical} as the RL algorithm for training the verifier. Given a verification instance $(q_i, \tau_i)$, we sample a group of $G$ verification trajectories $\{\mathbf{o}_i\}_{i=1}^G$ from the current policy $\pi_\theta$. Each trajectory $\mathbf{o}_i = (o_{i}^1, \ldots, o_{i}^{|\mathbf{o}_i|})$ consists of reasoning tokens, search queries, retrieved documents, and a final judgment. The objective is:
{
\small
\begin{equation}
    \frac{1}{G} \sum_{i=1}^{G} \sum_{t=1}^{|\mathbf{o}_i|} \left\{ \min\left[r_{i}^{t} \hat{A}_{i}^{t}, \; \text{clip}\left( r_{i}^{t}, 1-\epsilon_{l}, 1+\epsilon_{h} \right) \hat{A}_{i}^{t} \right] \right\},
    \label{eq:drgrpo}
\end{equation}
}%
where $r_{i}^{t}
= \frac{\pi_\theta(o_{i}^{t} \mid \mathbf{q}, \mathbf{o}_{i}^{<t})}
       {\pi_{\theta_{\text{old}}}(o_{i}^{t} \mid \mathbf{q}, \mathbf{o}_{i}^{<t})}$, $\mathbf{q} = (q, \tau)$ denotes the input prompt containing the question and reasoning trace, $\mathbf{o}_{i}^{<t}$ represents previously generated tokens, and $\epsilon_l$ and $\epsilon_h$ are the clipping parameters. The advantage term $\hat{A}_{i}^{t}$ is defined as:
\[
    \hat{A}_{i}^{t} = R(\mathbf{q}, \mathbf{o}_i) - \text{mean}\left(\{R(\mathbf{q}, \mathbf{o}_1), \ldots, R(\mathbf{q}, \mathbf{o}_G)\}\right).
    \label{eq:drgrpo_advantage}
\]

\paragraph{Reward Designs.}

To facilitate multi-turn RL with tool execution, we design a structured reward covering two complementary objectives, following prior practices ~\cite{jin2025searchr1trainingllmsreason}:

(i) \textit{Correctness Reward ($R_c$)}: This component measures whether the verifier's judgment aligns with the ground-truth label. Let $\mathbf{q} = (q, \tau)$ denote the verification prompt and $\ell \in \{0, 1\}$ the ground-truth label. We define:
\[
    R_c = \mathbb{1}\big(\texttt{extract}(\mathbf{o}) = \ell\big),
\]
where $\mathbb{1}(\cdot)$ is the indicator function and $\texttt{extract}(\mathbf{o})$ parses the final judgment from the \texttt{<answer>} tags in the generated trajectory $\mathbf{o}$. Intuitively, $R_c = 1$ if the verifier's decision is correct, and $R_c = 0$ otherwise.

(ii) \textit{Format Reward ($R_f$)}: To ensure reliable tool use and structured outputs, the verifier is required to adhere to a predefined format. Specifically, reasoning steps must be enclosed within \texttt{<think>} tags, search queries within \texttt{<search>} tags, and the final judgment within \texttt{<answer>} tags. To discourage degenerate outputs, we further penalize excessive tag usage. Specifically, $R_f = 1$ if the output satisfies all formatting constraints and contains no more than 10 \texttt{<answer>} tag pairs; $R_f = 0.25$ if the output is correct but exhibits tag overflow; and $R_f = 0$ otherwise.

The final reward $R$ is defined as the product of the two components:
\[
    R = R_c \times R_f.
    \label{eq:reward}
\]

\subsection{Training Strategies}
\label{sec:training stra}
\paragraph{Adaptive Curriculum Formulation.}

A central challenge in RL for verification is ensuring that training data remains appropriately calibrated to the evolving capability of the model. Instances that are either trivially easy or impossibly difficult yields minimal learning signal, as the resulting policy gradients approach zero. To address this issue, we adopt a model-aware curriculum formulation mechanism that dynamically adapts the task distribution at each training iteration.

Concretely, before each iteration $t$, we perform online filtering on the sampled batch $\mathcal{B}_t$ to construct an effective training set $\mathcal{D}_t$:
\[
    \mathcal{D}_t = \{ (q, \tau, \ell) \in \mathcal{B}_t : \exists\, g, g' \in \{1, \ldots, G\} \\ 
    \text{ s.t. } r^{(g)} \neq r^{(g')} \}.
\]
Here, for each candidate instance $(q, \tau, \ell) \in \mathcal{B}_t$, we sample $G$ verification trajectories $\{o^{(g)}\}_{g=1}^{G}$ from the current policy $\pi_{\theta_t}$. We then compute the corresponding rewards $\{r^{(g)}\}_{g=1}^{G}$. Finally, we retain only instances if any two rewards are different, i.e., reward variance is non-zero~\cite{khatri2025artscalingreinforcementlearning}.

This criterion eliminates prompts where the model either consistently succeeds or consistently fails across all sampled trajectories. By filtering these zero-gradient instances, optimization is focused on decision-boundary cases where the verifier exhibits uncertainty.

To maintain a fixed training budget per iteration, we iteratively resample additional instances from the labeled pool $\mathcal{B}$ and apply the same filtering criterion until $|\mathcal{D}_t| = 20K$. This dynamic curriculum evolves naturally across iterations as the verifier improves, eliminating the need for manually designed difficulty schedules.

\begin{algorithm}[t]
\caption{Iterative Training of Tool-Integrated Medical Reasoning Verifier}
\label{alg:med_tiv}
\begin{algorithmic}[1]
\REQUIRE Base verifier $\pi_{\theta_0}$, labeled dataset pool $\mathcal{D} = \{(q_i, \tau_i, \ell_i)\}_{i=1}^{N}$, maximum iterations $T_{\max}$, batch size $B$, group size $G$, search engine $\mathcal{E}$
\ENSURE Trained verifier $\pi_{\theta^*}$

\FOR{$t = 1$ to $T_{\max}$}

    \STATE \textcolor{blue}{\(\triangleright\) Sample labeled batch}
    \STATE $\mathcal{B}_t \leftarrow \textsc{SampleBatch}(\mathcal{D}, B)$
    \STATE $\mathcal{D}_t \leftarrow \emptyset$

    \STATE \textcolor{blue}{\(\triangleright\) Curriculum formulation}
    \FOR{each $(q, \tau, \ell) \in \mathcal{B}_t$}
        \STATE Sample verification trajectories:\\
        \quad $\{\hat{\ell}^{(g)}\}_{g=1}^{G} \sim \pi_{\theta_t}(\cdot \mid q, \tau, \mathcal{E})$
        \STATE Compute rewards within group:\\
        \quad $r^{(g)} \leftarrow \mathbb{1}[\hat{\ell}^{(g)} = \ell]$, for $g \in 1, \ldots, G$
        \IF{$\exists\, g \neq g'$ such that $r^{(g)} \neq r^{(g')}$}
            \STATE Add $(q, \tau, \ell)$ to curriculum set $\mathcal{D}_t$
        \ENDIF
    \ENDFOR

    \STATE \textcolor{blue}{\(\triangleright\) RL optimization on curriculum data}
    \STATE $\pi_{\theta_{t+1}} \leftarrow \textsc{Dr.GRPO}(\pi_{\theta_t}, \mathcal{D}_t, \mathcal{E})$
\ENDFOR
\STATE \textbf{Return} $\pi_{\theta_{T_{\max}}}$
\end{algorithmic}
\end{algorithm}

\paragraph{Iterative Training via Self-Bootstrapping.}

We adopt an iterative training approach that progressively improves verification capabilities through multiple rounds of RL. Unlike prior work that alternates between rejection sampling, supervised fine-tuning (SFT), and RL \citep{xu2025incentivizingagenticreasoningllm}, our approach operates entirely through iterative RL, following the RL-Zero paradigm where the model reinforces its verification capabilities without requiring dense turn-level expert demonstrations for cold start.

Starting from the base model $\pi_{\theta_0}$, we perform $T_{\max}$ iterations. Each iteration consists of three stages:
\begin{align*}
\mathcal{B}_t &\leftarrow \textsc{SampleBatch}(\mathcal{D}, B), \\
\mathcal{D}_t &\leftarrow \textsc{Filter}(\mathcal{B}_t, \pi_{\theta_t}), \\
\pi_{\theta_{t+1}} &\leftarrow \textsc{RL}(\pi_{\theta_t}, \mathcal{D}_t).
\end{align*}
Each iteration draws a fresh batch $\mathcal{B}_t$ from the annotated pool $\mathcal{D}$ with trace-level labels, ensuring a balanced distribution of correct and incorrect reasoning traces. The curriculum filtering then constructs the training set $\mathcal{D}_t$ as described above, and RL optimization updates the policy based on the structured reward.

The key insight underlying this iterative approach is the co-evolution of model capability and training distribution. As the verifier improves, the filtering mechanism automatically removes instances that have become too easy, while the fresh sampling introduces new challenging cases. This creates a self-bootstrapping cycle: stronger models encounter harder verification tasks, which in turn drive further improvements. Since the trace-level correctness reward is deterministic and unambiguous, this self-bootstrapping process converges reliably without the instabilities that can arise from noisy synthetic step-level labels. We summarize the overall training procedure in Algorithm~\ref{alg:med_tiv}.

\section{Experiments}
\subsection{Experimental Setup}

\paragraph{Evaluation benchmarks.}

We evaluated \method on four open-source medical question-answering benchmarks: MedQA~\citep{jin2020diseasedoespatienthave}, MedMCQA~\citep{pal2022medmcqalargescalemultisubject}, MMLU-Med~\citep{hendrycks2021measuringmassivemultitasklanguage}, and MedXpertQA~\citep{zuo2025medxpertqabenchmarkingexpertlevelmedical}, using accuracy as the evaluation metric. These benchmarks collectively assess the verifier's ability to distinguish correct from erroneous reasoning across varying difficulty levels and medical subdomains. Detailed descriptions of benchmarks are in Appendix~\ref{sec:benchmarks}.

\paragraph{Implementation details.}

We trained verifiers using two light-weight backbone models: Llama3.1-8B and Qwen2.5-7B, with Llama3.1-8B as the default for results reporting. All training was conducted using the VeRL-Tool framework~\citep{jiang2025verltoolholisticagenticreinforcement}. Detailed hyperparameters are shown in Appendix \ref{sec:hyperparameters}. All experiments were conducted on 4 NVIDIA H100 GPUs with 80GB of memory. Due to computational constraints, we limit the maximum number of RL iterations to $T_{\max} = 2$ and we set the group size for curriculum formulation (Section~\ref{sec:training stra}) to $G=8$. 

For inference, we used the default sampling hyperparameters for all models. In reward-guided search experiments, unless otherwise specified, we used Qwen2.5-7B as the frozen generator and sampled up to 32 candidate reasoning traces per question. We applied Hard-Weighted Self-Consistency as the default test-time search strategy.

\paragraph{Baselines.}
We compared \method against two groups of baselines. 1): \textit{Off-the-shelf LLMs}: GPT-4o-mini~\cite{openai2024gpt4ocard}, Gemini-2.0-Flash, DeepSeek-R1 series~\cite{Guo_2025}, Qwen2.5 series~\cite{qwen2025qwen25technicalreport}, Llama3.1~\cite{grattafiori2024llama3herdmodels}, AlphaMed~\cite{liu2025distillationpushinglimitsmedical},  UltraMedical~\cite{zhang2024ultramedicalbuildingspecializedgeneralists}, and HuatuoGPT-o1~\cite{chen2024huatuogpto1medicalcomplexreasoning}. 2): \textit{Medical domain-specialized Reward Models}:
MedS$^3$~\cite{jiang2025meds3medicalslowthinking} and Med-PRM~\cite{yun2025medprmmedicalreasoningmodels}. Detailed descriptions of each reward model baseline are shown in Appendix~\ref{sec:baseline_setup}.

\begin{table*}[!ht]
    \centering
    \caption{Main evaluation results on medical reasoning benchmarks. We report accuracy (\%) on MedQA, MedMCQA, MMLU-Med, and MedXpertQA. 
    \textbf{Bold} numbers indicate the best results among the reward model group. \cmark: Verifier supports external tools for judging;
    \xmark: Verifier does not support external tools for judging.}
    % \small
    \begin{tblr}{
        colspec = {lccccccccc},
        row{1} = {bg=gray!25},
        row{3,4,6,7,8,10,11,12,14,15,16,18,19} = {bg=gray!10},
        row{20} = {bg=LightCyan},
        hline{1,2,Z} = {black},
        hline{5,9,13,17} = {gray},
        cells = {font = \small},
    }
    % \toprule
    \textbf{Baselines}
    & \textbf{\tool}
    & \textbf{$|$Train$|$}
    & \textbf{Size}
    & \textbf{MedQA}
    & \textbf{MedMCQA}
    & \textbf{MMLU-Med}
    & \textbf{MedXpertQA}
    & \textbf{Avg.} \\
    % \midrule

    \textit{Proprietary Models} & & & & & & & & \\
    %\midrule
    \quad GPT-4o-mini
        & -
        & - & -
        & 79.03 & 68.20 & 87.79 & 17.84 & 63.22 \\
    \quad Gemini-2.0-Flash
        & -
        & - & -
        & 87.51 & 72.60 & 92.01 & 20.57 & 68.17 \\

    %\midrule
    \textit{General Reasoning Models} & & & & & & & & \\
    %\midrule
    \quad DeepSeek-R1
        & -
        & - & 671B
        & 90.34 & 78.80 & 94.40 & 37.76 & 75.33 \\
    \quad R1-Distill-Qwen
        & -
        & - & 7B
        & 24.82 & 36.40 & 47.47 & 7.43 & 29.03 \\
    \quad R1-Distill-Llama
        & -
        & - & 8B
        & 34.96 & 43.60 & 64.19 & 5.35 & 37.03 \\

    %\midrule
    \textit{General Non-reasoning Models} & & & & & & & & \\
    %\midrule
    \quad Qwen2.5
        & -
        & - & 32B
        & 73.21 & 64.83 & 84.94 & 13.87 & 59.21 \\
    \quad Qwen2.5
        & -
        & - & 7B
        & 60.96 & 56.56 & 76.96 & 12.15 & 51.66 \\
    \quad Llama3.1
        & -
        & - & 8B
        & 70.93 & 61.60 & 78.97 & 13.02 & 56.13 \\

    %\midrule
    \textit{Medical Reasoning Models} & & & & & & & & \\
    %\midrule
    \quad AlphaMed
        & -
        & - & 7B
        & 71.01 & 61.46 & 81.16 & 19.16 & 58.20 \\
    \quad UltraMedical
        & -
        & - & 8B
        & 72.66 & 62.60 & 79.61 & 15.25 & 57.53 \\
    \quad HuatuoGPT-o1
        & -
        & - & 8B
        & 72.19 & 63.60 & 75.30 & 16.84 & 56.98 \\

    %\midrule
    \textit{Medical Reward Models} & & & & & & & & \\
    %\midrule
    \quad MedS$^3$
        & \xmark
        & 225k & 7B
        & 64.89 & 58.91 & 80.53 & 12.90 & 54.31 \\
    \quad Med-PRM
        & \cmark
        & 111k & 7B
        & 69.99 & 62.36 & 80.99 & 13.51 & 56.71 \\
    \quad \textbf{$\method$ (Ours)}
        & \cmark
        & 20k & 7B
        & \textbf{75.26} & \textbf{64.70} & \textbf{85.58} & \textbf{16.04} & \textbf{60.40} \\

    % \bottomrule
    \end{tblr}

    \vspace{2mm}
    \label{tab:main_results}
\end{table*}

\paragraph{Test-Time Search Strategies.}

We evaluated three test-time search strategies that leverage \method to improve the reasoning performance of frozen generators. Given a reasoning trace $\tau = (s_1, s_2, \dots, s_K)$ with $K$ steps, our verifier assigns a confidence score $r_\tau \in [0, 1]$ for the entire trace, defined as the softmax probability of the \texttt{1} token over the logits of both \texttt{1} and \texttt{0} tokens.

\begin{itemize}[leftmargin=*,nosep]
    \item \textit{Best-of-N.} Given a question $q$, we sampled $N$ candidate traces $\{\tau^{(j)}\}_{j=1}^N$ from the generator and selected the trace with the highest verifier confidence score:
\[
    \hat{\tau} = \arg\max_{\tau^{(j)}} r_{\tau^{(j)}}.
\]
    \item \textit{Hard-Weighted Self-Consistency.} We first filtered traces by the verifier's binary judgment, keeping only those labeled correct ($V_\theta(q, \tau) = 1$). Among the filtered traces, we applied majority voting to determine the final answer:
\[
    \hat{a} = \arg\max_{a} \sum_{j=1}^{N} \mathbb{1}\big[V_\theta(q, \tau^{(j)}) = 1\big] \cdot \mathbb{1}\big[\text{ans}(\tau^{(j)}) = a\big].
\]
    \item \textit{Soft-Weighted Self-Consistency.} Instead of binary filtering, we weighted each trace's vote by the verifier's confidence score:
\[
    \hat{a} = \arg\max_{a} \sum_{j=1}^{N} r_{\tau^{(j)}} \cdot \mathbb{1}\big[\text{ans}(\tau^{(j)}) = a\big].
\]
\end{itemize}

\subsection{Main Results}

Table~\ref{tab:main_results} presents the main results on four medical reasoning benchmarks. Models trained with \method consistently outperform existing baselines across all benchmarks. Specifically, under guided-search using a \method-trained verifier, Qwen2.5-7B attains accuracies of 75.26\% on MedQA, 64.70\% on MedMCQA, 85.58\% on MMLU-Med, and 16.04\% on MedXpertQA, yielding an average accuracy of 60.40\%. 
Notably, \method enables this 7B generator to rival substantially larger models, even surpassing the base performance of Qwen2.5-32B despite using a generator that is approximately $\mathbf{4\times}$ smaller. Compared to domain-specialized medical reasoning models of similar scale, \method outperforms HuatuoGPT-o1-8B and UltraMedical-8B by 3.07\% and 2.60\% on MedQA, respectively, demonstrating the effectiveness of our tool-integrated verification. Case analysis in Appendix~\ref{app:case_ana} further illustrates how \method identifies subtle reasoning errors.

\subsection{Analysis}
\label{sec:analysis}

We conducted a series of ablation analyses to investigate six key research questions regarding our proposed framework. 
% \textbf{(RQ1)} Does $\method$ generalize across different generator models? 
% \textbf{(RQ2)} How do different test-time search strategies compare under $\method$?
% \textbf{(RQ3)} Can $\method$ reduce the sampling budget required to achieve state-of-the-art performance compared to existing baselines?
% \textbf{(RQ4)} What is the impact of iterative training?
% \textbf{(RQ5)} Does $\method$ generalize across different base models? 
% We address each question through targeted experiments below.

\begin{table}[t]
    \centering
    \caption{Performance improvements from using $\method$ as a verifier on MedQA. For each generator model, the first row indicates the accuracy over single sampled trace per question.}
    
    %\resizebox{\linewidth}{!}{%
    \begin{tblr}{
        colspec = {lc},
        row{1} = {bg=gray!25},
        row{3,4,5,6,8,9,10,11,13,14,15,16} = {bg=gray!10},
        row{6,11,16} = {bg=LightCyan},
        hline{1,2,Z} = {black},
        hline{7,12} = {gray},
        cells = {font = \small},
    }
    % \toprule
    \textbf{Models} & \textbf{MedQA} \\
    % \midrule
    \textit{Qwen2.5-7B} & 60.96 \\
    \quad + Self-Consistency & 66.38 (+5.42) \\
    \quad + Best-of-N (Med-TIV) & 72.35 (+11.39) \\
    \quad + Soft Weighted SC (Med-TIV) & 75.02 (+14.06) \\
    \quad + Hard Weighted SC (Med-TIV) & \textbf{75.26} (+14.30) \\
    % \midrule
    \textit{AlphaMed-7B} & 71.01 \\
    \quad + Self-Consistency & 74.23 (+3.22) \\
    \quad + Best-of-N (Med-TIV) & 75.02 (+4.01) \\
    \quad + Soft Weighted SC (Med-TIV) & 75.33 (+4.32) \\
    \quad + Hard Weighted SC (Med-TIV) & \textbf{75.65} (+4.64) \\
    % \midrule
    \textit{Qwen2.5-32B} & 73.21 \\
    \quad + Self-Consistency & 75.26 (+2.05) \\
    \quad + Best-of-N (Med-TIV) & 75.57 (+2.36) \\
    \quad + Soft Weighted SC (Med-TIV) & 75.96 (+2.75) \\
    \quad + Hard Weighted SC (Med-TIV) & \textbf{75.96} (+2.75) \\
    % \bottomrule
    \end{tblr}%
    %}
    \label{tab:guided_search}
\end{table}

\begin{figure*}[t]
    \centering
    \begin{subfigure}[b]{1.0\textwidth}
        \centering
        \includegraphics[width=\linewidth]{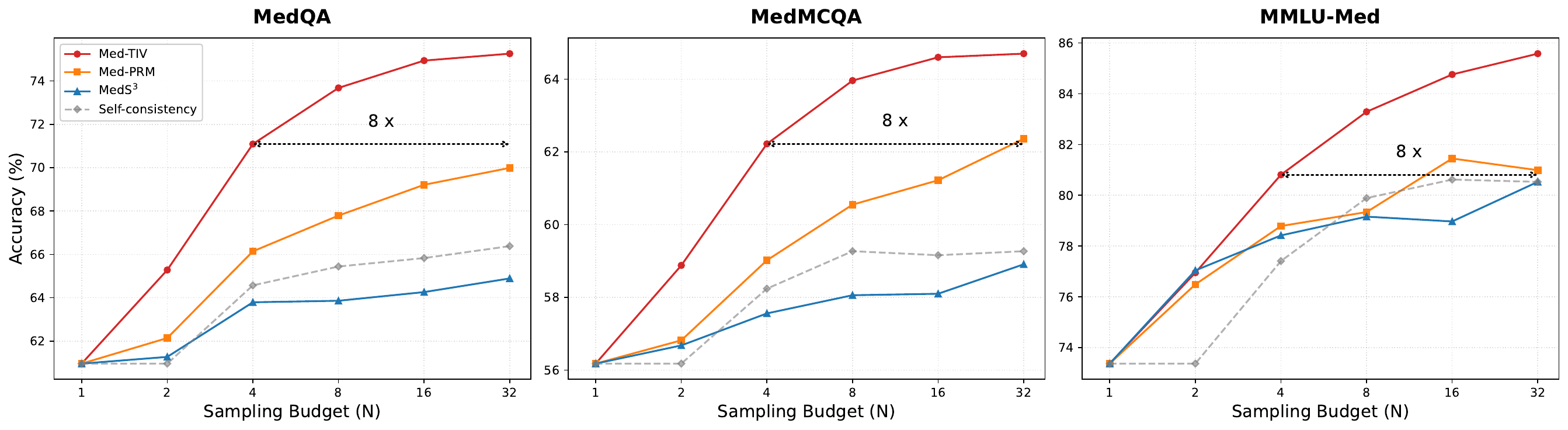}
    \end{subfigure}
    \caption{Test-time scaling analysis across three medical reasoning benchmarks. Each plot shows accuracy versus sampling budget $N \in \{1, 2, 4, 8, 16, 32\}$ for four baselines. $\method$ consistently outperforms baselines across all sampling budgets and benchmarks.}
    \label{fig:scaling}
\end{figure*}

\paragraph{Q1: Does \method generalize across different generator models?}
To evaluate the generalizability of the trained verifier, we applied \method to guide test-time search across generator models of varying sizes and capabilities. As shown in Table~\ref{tab:guided_search}, when using Qwen2.5-7B as the generator, Hard-Weighted Self-Consistency yields a relative improvement of 23.5\% over the base model's single-sample accuracy, substantially outperforming the 12.2\% gain achieved by standard Self-Consistency. Notably, the domain-specialized AlphaMed-7B model also benefits from verifier guidance with a 6.5\% relative improvement, indicating that our verifier provides complementary verification capabilities beyond domain-specific fine-tuning. The improvements extend to larger models as well: Qwen2.5-32B achieves a 3.8\% relative gain during test-time search, demonstrating that a light-weight 8B verifier can effectively guide models that are significantly larger than itself. This cross-scale generalization suggests that \method learns transferable verification patterns rather than overfitting to specific generator characteristics.

\paragraph{Q2: How do different test-time search strategies compare under \method?}

We then systematically compare different test-time search strategies under verifier guidance to identify the most effective approach for leveraging verification signals. As shown in Table~\ref{tab:guided_search}, Hard-Weighted Self-Consistency consistently achieves the highest accuracy across all generators, followed by Soft-Weighted Self-Consistency and Best-of-$N$ selection. On Qwen2.5-7B, Hard-Weighted Self-Consistency outperforms Best-of-$N$ by 3\% absolute accuracy, suggesting that majority voting among verified traces provides more robust answer selection than simply choosing the highest-confidence individual trace.

\paragraph{Q3: Can \method reduce the sampling budget required to achieve state-of-the-art performance compared to existing baselines?}

Next, we investigated how verification performance scales with sampling budget, a critical consideration for deployment under varying computational constraints. 
%The scaling analysis reveals a striking efficiency advantage for $\method$ over existing medical reward models. 
As shown in Figure~\ref{fig:scaling}, \method achieves substantial efficiency advantage over existing medical reward models across all three benchmarks.
In particular, \method matches the performance of baselines using only 4 samples, whereas the baselines require 32 samples, representing an $\mathbf{8\times}$ reduction in sampling budget. On MedQA, \method achieves 72.1\% accuracy at $N=4$, while Med-PRM requires the full $N=32$ budget to reach 70.0\% accuracy. Since inference cost scales approximately linearly with the number of sampled traces, this translates to equivalent performance at one-eighth the generator inference cost in practical deployment settings.  

% Moreover, the scaling curves reveal diminishing returns beyond 16 samples for \method, suggesting that the verifier effectively identifies high-quality traces early in the sampling process. In contrast, baseline methods continue to benefit from additional samples, but with slower rate.

\begin{figure}[t]
    \centering
    \includegraphics[width=0.95\linewidth]{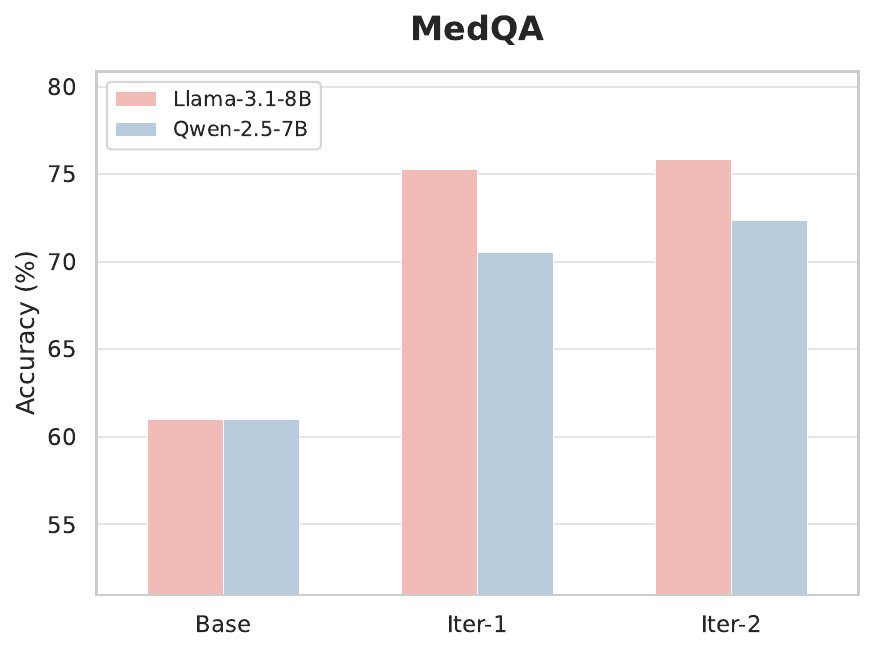}
    \caption{Ablation on base model selection and training iterations.}
    \label{fig:medqa_ablation}
\end{figure}

\paragraph{Q4: Does \method generalize across different base models?}

To assess the generality of our proposed framework, we compared verification performance using two distinct verifier backbones: Llama3.1-8B and Qwen2.5-7B. As shown in Figure~\ref{fig:medqa_ablation}, both backbones achieve strong performance after two training iterations. Llama3.1-8B consistently outperforms Qwen2.5-7B by approximately 3.5\% absolute accuracy on MedQA, achieving 75.86\% versus 72.35\% after 2 iterations of training. The parallel performance gains observed across both models indicate that \method is agnostic to backbone architectures.

\paragraph{Q5: What is the impact of iterative training?}

Figure~\ref{fig:medqa_ablation} presents ablation results examining the impact of iterative training with adaptive curriculum formulation. Llama3.1-8B improves from 60.96\% to 75.26\% after iteration 1, with marginal gains to 75.86\% at iteration 2. Qwen2.5-7B follows a similar pattern, reaching 72.35\% after two iterations. The rapid convergence suggest that the majority of verification capability is acquired in the first round, with subsequent iterations refining boundary cases.

\begin{wraptable}{r}{0.32\textwidth} % Increased slightly to fit deltas
\vspace{-1em}
\centering
\caption{Ablation on RL and tool integration.}
\label{tab:tool_ablation}
\begin{tblr}{
    colspec = {l r}, % 'r' column aligns numbers/deltas better
    row{1} = {bg=gray!25},
    row{4,7} = {bg=gray!10}, % Highlights the Base Generator names
    cells = {font=\small},
    hline{1,2,8} = {black},
    hline{5} = {gray}, % Separator between model blocks
}
\textbf{Models} & \textbf{MedQA} \\
Qwen2.5-7B & 60.96 \\
\quad + Med-TIV (RL) & 69.60 \\
\quad + Med-TIV (RL + Tool) & \textbf{70.54} \\
AlphaMed-7B & 71.01 \\
\quad + Med-TIV (RL) & 76.12 \\
\quad + Med-TIV (RL + Tool) & \textbf{77.14} \\
\end{tblr}
\vspace{-1em}
\end{wraptable}

\paragraph{Q6: How does RL and tool integration impact verification performance?}

Table~\ref{tab:tool_ablation} highlights the dual benefits of our framework across two generators. RL training drives the primary gain, boosting MedQA accuracy of Qwen2.5-7B by 8.64\%, confirming that the verifier effectively internalizes reasoning patterns. Tool integration provides a critical secondary boost, further elevating accuracy to 70.54\%. A similar cumulative trend is observed with AlphaMed-7B. This demonstrates that while RL anchors logical verification, dynamic retrieval is essential for resolving knowledge-intensive boundary cases beyond the model's parametric memory.

% These consistent improvements across models confirm the effectiveness of  adaptive curriculum formulation. Moreover, the similarly magnitude of gains from the second iteration suggests that our iterative training approach is robust and does not depend strongly on the choice of base model.

\section{Related Work}
\paragraph{Medical Reasoning Models.}
The application of large language models to medical reasoning has attracted considerable attention. Early efforts focused on domain-adaptive pretraining and instruction tuning on medical corpora ~\citep{wu2023pmcllamabuildingopensourcelanguage, singhal2025expert, chen2023meditron70bscalingmedicalpretraining}. More recent work has explored reasoning-enhanced medical models. HuatuoGPT-o1 ~\citep{chen2024huatuogpto1medicalcomplexreasoning} incorporates chain-of-thought reasoning with verification mechanisms, and UltraMedical ~\citep{zhang2024ultramedicalbuildingspecializedgeneralists} combines high-quality instruction data with preference optimization. AlphaMed~\citep{liu2025distillationpushinglimitsmedical} employs RL to improve medical reasoning capabilities. Despite these advances, most existing approaches focus on improving the generator model itself, whereas our work addresses the complementary problem of training a plug-and-play verifier that can improve any frozen generator through test-time search.

\paragraph{Tool-Assisted Reward and Judge Models.}
Standard LLM-based judges typically function as passive scorers limited by parametric knowledge. Recent work addresses this through agentic reward modeling, equipping verifiers with executable tools. Themis~\citep{li2024toolaugmentedrewardmodeling} established the foundational framework by enabling access to calculators, search engines, and knowledge bases through structured tool-calling traces. TIR-Judge~\cite{xu2025incentivizingagenticreasoningllm} advanced this paradigm in the general domain by integrating code execution to judge paired responses. TIM-PRM~\citep{kuang2025timprmverifyingmultimodalreasoning} introduced independent tool queries for multi-modal verification to eliminate confirmation bias. The concept has further expanded to the Agent-as-a-Judge paradigm ~\cite{you2026agentasajudge}, which employs dynamic planning, tool augmentation and multi-agent coordination to decompose complex evaluation tasks. Our work instantiates this agentic paradigm within the medical domain, moving beyond static retrieval to iterative, evidence-grounded clinical verification.

\section{Conclusion}

We presented \method, an agentic RL framework for medical reasoning verification. Our approach addresses key limitations of existing medical reward models by offering explicit critique traces and enabling dynamic knowledge retrieval during verification. Empirical evaluations across four medical reasoning benchmarks demonstrate that \method substantially outperforms prior approaches. More broadly, \method introduces a general paradigm for training tool-augmented verifiers that can be extended to other high-stakes domains requiring evidence-grounded evaluation.

\section*{Impact Statement}
This paper introduces research aimed at improving the reliability of large language models for medical reasoning tasks. We believe our work contributes positively to the development of trustworthy medical AI systems by providing mechanisms to verify reasoning correctness before clinical deployment. \method holds potential to enhance the safety of LLM-assisted clinical decision support by reducing erroneous reasoning outputs through systematic verification. By grounding judgments in retrieved medical evidence, our approach offers improved transparency compared to opaque scalar reward models, enabling practitioners to better understand and audit verification decisions. The efficiency gains demonstrated by $\method$ could democratize access to reliable medical reasoning verification, making robust verification feasible even in resource-constrained settings.

% However, we acknowledge important limitations and potential risks. $\method$ cannot fully address biases present in underlying medical corpora, which may reflect historical inequities in clinical research and practice. The system should not be deployed as a standalone diagnostic tool but rather as one component within broader clinical workflows that include human oversight. Additionally, over-reliance on automated verification could potentially reduce critical engagement by healthcare professionals. We recommend that any clinical deployment of the systems undergo rigorous validation studies and maintain appropriate human-in-the-loop safeguards to ensure patient safety.

\section*{Acknowledgement}
This study was supported by the National Institutes of Health awards UL1TR001857, U24TR004111, R01LM014588, and R01LM014306. The sponsors had no role in study design, data collection, analysis, interpretation, report writing, or decision to submit the paper for publication.

\nocite{langley00}

\bibliography{example_paper}
\bibliographystyle{icml2026}

%%%%%%%%%%%%%%%%%%%%%%%%%%%%%%%%%%%%%%%%%%%%%%%%%%%%%%%%%%%%%%%%%%%%%%%%%%%%%%%
%%%%%%%%%%%%%%%%%%%%%%%%%%%%%%%%%%%%%%%%%%%%%%%%%%%%%%%%%%%%%%%%%%%%%%%%%%%%%%%
% APPENDIX
%%%%%%%%%%%%%%%%%%%%%%%%%%%%%%%%%%%%%%%%%%%%%%%%%%%%%%%%%%%%%%%%%%%%%%%%%%%%%%%
%%%%%%%%%%%%%%%%%%%%%%%%%%%%%%%%%%%%%%%%%%%%%%%%%%%%%%%%%%%%%%%%%%%%%%%%%%%%%%%
\newpage
\appendix
\onecolumn
\section{Limitation}\label{sec:limitations}
While $\method$ demonstrates substantial improvements over existing medical reasoning verification approaches, several limitations warrant discussion and suggest directions for future research.

\paragraph{Process Supervision.} Our current training paradigm relies solely on trace-level outcome rewards, providing no supervision on intermediate verification behaviors such as when to search, what queries to formulate, or how to integrate retrieved evidence. While this design eliminates the need for costly step-level annotations, it may lead to suboptimal search patterns or redundant retrieval operations. Future work could explore supervision for the verification task itself, or leverage techniques such as search behavior cloning from stronger models to provide denser optimization signals.

\paragraph{Retrieval Corpus Coverage.} $\method$'s verification accuracy is inherently bounded by the coverage and quality of the underlying medical corpus. Our retrieval system indexes documents from PubMed abstracts and medical textbooks, which provides broad coverage of established medical knowledge but may lack recent findings, rare disease information, or region-specific clinical guidelines. Verification of reasoning traces involving cutting-edge treatments or highly specialized subspecialties may be limited by corpus gaps.

\paragraph{Language and Domain Scope.} All training and evaluation are conducted on English-language medical reasoning benchmarks. The generalization of Med-TIV to multilingual medical content or non-Western medical traditions remains unexplored. Additionally, while our benchmarks span multiple medical subdomains, certain specialized areas such as genomics, radiology interpretation, and surgical planning may require domain-adapted retrieval corpora for optimal verification performance.

\section{Additional Implementation Details}\label{sec:imp_detail}
\subsection{Hyperparameter Settings}
\label{sec:hyperparameters}

Table~\ref{tab:hyperparameters} provides comprehensive hyperparameter configurations for Med-TIV training across both iterations. We maintain mostly consistent settings between iterations to isolate the effect of iterative training from hyperparameter tuning.

\begin{table}[h]
\centering
\caption{Hyperparameter configurations for Med-TIV training across iterations.}
\label{tab:hyperparameters}
\begin{tabular}{lcc}
\toprule
\textbf{Hyperparameters} & \textbf{Iteration 1} & \textbf{Iteration 2} \\
\midrule
RL Algorithm & Dr.GRPO & Dr.GRPO \\
Clip ratio (low / high) & 0.2 / 0.3 & 0.2 / 0.3 \\
Learning rate & 1e-6 & 1e-6 \\
Warmup steps & 10 & 10 \\
Training epochs & 5 & 5 \\
Global batch size & 256 & 256 \\
Mini-batch size & 256 & 256 \\
Group size ($G$) & 5 & 8 \\
Rollout sampling temperature & 1.0 & 1.0 \\
Rollout top-p & 0.95 & 0.95 \\
Curriculum filtering & Enabled & Enabled \\
\bottomrule
\end{tabular}
\end{table}

\subsection{Retrieval Setup}\label{sec:retrieval_setup}
We construct our retrieval infrastructure using a dense retrieval architecture optimized for medical domain queries. The corpus is derived from the MedRAG~\cite{zhao2025medragenhancingretrievalaugmentedgeneration} collection, specifically combining the PubMed and Textbooks subcorpora into a unified index. The PubMed subset contains approximately 23.9 million biomedical abstracts covering research publications, while the Textbooks subset includes content from standard medical textbooks spanning clinical medicine, pharmacology, pathology, and related disciplines. After deduplication and quality filtering, the combined corpus contains approximately 24 million snippets. 

We employ MedCPT~\cite{Jin_2023} as our dense retrieval encoder, specifically the query encoder variant for encoding search queries and article encoder for encoding corpus snippets. Document embeddings are pre-computed and stored in a FAISS index using the Flat configuration for maximum retrieval accuracy, distributed across multiple GPUs using FAISS's GPU sharding capability to enable parallel similarity search. For each search query, we retrieve the top-3 most relevant documents for both training and inference.

\subsection{Baseline Setup}\label{sec:baseline_setup}

We describe the configuration of reward model baselines used in our experiments. For Med-PRM, which employs static retrieval-augmented generation, we equip it with the same retrieval corpus, encoder, and top-k setting as our framework to ensure a controlled comparison. MedS$^3$ does not support external tool invocation and is therefore evaluated without retrieval augmentation. For confidence score extraction and inference hyperparameter settings, we follow the configurations specified in each baseline's original publication.

\subsection{Prompt Template}
\label{sec:prompt}

We design a structured prompt template that guides the verifier through systematic reasoning with explicit tool invocation syntax. The complete prompt is shown in Table~\ref{tab:prompt}.

\begin{table}[t]
\centering
\caption{Prompt template.}
\label{tab:prompt}
\begin{tcolorbox}[
    width=\linewidth,
    colback=gray!8,        % light gray background
    colframe=black,        % black border
    boxrule=0.8pt,         % border thickness
    arc=0mm,               % sharp corners
    left=8pt,
    right=8pt,
    top=6pt,
    bottom=6pt
]

\textbf{User Prompt:}

You are a reasoning validator for medical problems. Your task is to think step by step and evaluate whether the given reasoning trace of a medical problem contains errors.

First, you must always perform a step-by-step analysis to examine the entire reasoning process. Then, based on your analysis, you will make a definitive judgment.

- Use 1 if the reasoning trace is free of errors.

- Use 0 if the reasoning trace contains one or more errors.

Output Instruction:

You must conduct your step-by-step analysis inside \texttt{<think>} and \texttt{</think>} first every time you get new information. After reasoning, if you find you lack some knowledge, you can call a search engine by \texttt{<search>} query \texttt{</search>} and it will return the top searched results between \texttt{<information>} and \texttt{</information>}. You can search as many times as you want. If you find no further external knowledge needed, you can directly provide the answer inside \texttt{<answer>} and \texttt{</answer>}, without detailed illustrations.

\begin{verbatim}
Medical Problem:
{The full Medical Problem on one or more lines.}

Reasoning Trace:
{The full Reasoning Trace on one or more lines.}
\end{verbatim}

\end{tcolorbox}
\end{table}

\section{Benchmarks and Baselines}
\label{sec:benchmarks_baselines}

\subsection{Benchmarks}
\label{sec:benchmarks}

We evaluate Med-TIV on four established medical reasoning benchmarks that collectively assess verification capability across varying difficulty levels and medical subdomains.

\begin{itemize}[leftmargin=*,nosep]
    \item \textbf{MedQA}~\citep{jin2020diseasedoespatienthave}: A dataset of multiple-choice questions derived from the United States Medical Licensing Examination (USMLE), designed to evaluate clinical reasoning and medical knowledge integration across diverse specialties.
    
    \item \textbf{MedMCQA}~\citep{pal2022medmcqalargescalemultisubject}: A large-scale multi-subject benchmark sourced from Indian medical entrance examinations (AIIMS and NEET-PG), covering 21 medical subjects with emphasis on factual knowledge and clinical application.
    
    \item \textbf{MMLU-Med}~\citep{hendrycks2021measuringmassivemultitasklanguage}: An aggregation of medical-related subsets from the Massive Multitask Language Understanding benchmark, encompassing anatomy, clinical knowledge, college biology, college medicine, medical genetics, and professional medicine.
    
    \item \textbf{MedXpertQA}~\citep{zuo2025medxpertqabenchmarkingexpertlevelmedical}: An expert-level benchmark featuring challenging questions that require multi-step clinical reasoning, differential diagnosis, and treatment planning at the level expected of practicing physicians.
\end{itemize}

\subsection{Baselines}
\label{sec:baselines}

We compare Med-TIV against comprehensive baselines spanning proprietary systems, general-purpose models, and domain-specialized approaches.

\paragraph{Proprietary Models.}
\begin{itemize}[leftmargin=*,nosep]
    \item \textbf{GPT-4o-mini}~\citep{openai2024gpt4ocard}: A compact variant of OpenAI's GPT-4o optimized for efficiency while maintaining strong reasoning capabilities across diverse tasks.
    
    \item \textbf{Gemini-2.0-Flash}: Google's efficient multimodal model designed for fast inference with competitive performance on knowledge-intensive benchmarks.
\end{itemize}

\paragraph{General Reasoning Models.}
\begin{itemize}[leftmargin=*,nosep]
    \item \textbf{DeepSeek-R1}~\citep{Guo_2025}: A 671B parameter reasoning model trained with RL, representing the current frontier of open-weight reasoning capabilities.
    
    \item \textbf{R1-Distill-Qwen / R1-Distill-Llama}: Distilled variants of DeepSeek-R1 at 7B and 8B scales respectively, designed to transfer reasoning capabilities to smaller architectures.
\end{itemize}

\paragraph{General Foundation Models.}
\begin{itemize}[leftmargin=*,nosep]
    \item \textbf{Qwen2.5}~\citep{qwen2025qwen25technicalreport}: A family of open-weight language models with strong multilingual and reasoning capabilities, evaluated at 7B and 32B parameter scales.
    
    \item \textbf{Llama3.1}~\citep{grattafiori2024llama3herdmodels}: Meta's open-source foundation model demonstrating competitive performance across diverse benchmarks, evaluated at the 8B scale.
\end{itemize}

\paragraph{Medical Domain Models.}
\begin{itemize}[leftmargin=*,nosep]
    \item \textbf{AlphaMed}~\citep{liu2025distillationpushinglimitsmedical}: A medical reasoning model that employs RL with rule-based rewards to enhance clinical reasoning without reliance on distillation from larger models.
    
    \item \textbf{UltraMedical}~\citep{zhang2024ultramedicalbuildingspecializedgeneralists}: A specialized medical model combining high-quality instruction tuning on curated biomedical corpora with preference optimization for improved clinical accuracy.
    
    \item \textbf{HuatuoGPT-o1}~\citep{chen2024huatuogpto1medicalcomplexreasoning}: A medical reasoning model incorporating chain-of-thought reasoning with internal verification mechanisms to improve diagnostic accuracy.
\end{itemize}

\paragraph{Medical Reward Models.}
\begin{itemize}[leftmargin=*,nosep]
    \item \textbf{MedS$^3$}~\citep{jiang2025meds3medicalslowthinking}: A self-evolved soft dual-sided process supervision framework for medical reasoning that generates training signals through iterative self-improvement without external annotations.
    
    \item \textbf{Med-PRM}~\citep{yun2025medprmmedicalreasoningmodels}: A process reward model for medical reasoning verification that provides step-level supervision using static retrieval-augmented generation with guideline-based verification.
\end{itemize}

\section{Case Analysis}\label{app:case_ana}

Table~\ref{tab:case} presents a complete verification example illustrating how a $\method$ trained verifier identifies reasoning errors through dynamic evidence retrieval. The case involves a patient with bladder cancer who develops ototoxicity following chemotherapy. The generator's reasoning trace incorrectly attributes the symptoms to taxanes based on their known association with ototoxicity, concluding with answer (B). However, the model retrieves evidence establishing that cisplatin—the standard neoadjuvant therapy for transitional cell carcinoma—is the causative agent, and its mechanism involves DNA cross-linking rather than microtubule hyperstabilization. Through iterative search and reasoning, verifier correctly identifies the error, demonstrating the value of tool augmentation for catching subtle medical reasoning mistakes.

\begin{table}[t]
\centering
\caption{Complete verification demonstration. Given a medical problem and a reasoning trace, the verifier retrieves relevant evidence and correctly identifies the reasoning error within the trace.}
\label{tab:case}
\begin{tcolorbox}[
    width=\linewidth,
    colback=gray!8,        % light gray background
    colframe=black,        % black border
    boxrule=0.8pt,         % border thickness
    arc=0mm,               % sharp corners
    left=8pt,
    right=8pt,
    top=6pt,
    bottom=6pt
]

\textbf{Medical Problem:}

Question: A 67-year-old man with transitional cell carcinoma of the bladder comes to the physician because of a 2-day history of ringing sensation in his ear. He received this first course of neoadjuvant chemotherapy 1 week ago. Pure tone audiometry shows a sensorineural hearing loss of 45 dB. The expected beneficial effect of the drug that caused this patient's symptoms is most likely due to which of the following actions?

Options: 
(A) Inhibition of proteasome (B) Hyperstabilization of microtubules 
(C) Generation of free radicals (D) Cross-linking of DNA
\vspace{8pt}

\textbf{Generator Reasoning Trace:}

The patient's presentation of ringing in the ear, known as tinnitus, and a sensorineural hearing loss of 45 dB, suggest ototoxicity. Ototoxicity is a well-known side effect of certain chemotherapy drugs. Neoadjuvant chemotherapy refers to the use of chemotherapy before other treatments such as surgery. The patient has received a course of neoadjuvant chemotherapy 1 week ago, which aligns with the recent onset of ototoxic symptoms. To determine the cause of ototoxicity from the provided options, we need to identify which of the options is likely to cause sensorineural hearing loss. Option (A), Inhibition of proteasome, is associated with drugs like bortezomib (Velcade), which are less commonly associated with ototoxicity. Option (B), Hyperstabilization of microtubules, is a known mechanism of taxanes (e.g., paclitaxel, docetaxel). Taxanes are known to cause ototoxicity, making this a potential candidate. Option (C), Generation of free radicals, is a general mechanism that can cause cellular damage but is not specifically known to be associated with ototoxicity. Option (D), Cross-linking of DNA, is a known mechanism of alkylating agents like cyclophosphamide. While alkylating agents can cause ototoxicity, this mechanism is not as commonly linked to ototoxicity as other options. Considering the options, the one most commonly associated with ototoxicity is hyperstabilization of microtubules. Therefore, the drug that most likely caused this patient's symptoms is one that hyperstabilizes microtubules. The answer is (B).

\vspace{8pt}
\textbf{Verifier Judgment:}

\texttt{<think>}
Let's check the common chemotherapeutic agents used for bladder cancer and their known side effects.
Transitional cell carcinoma of the bladder is often treated with cisplatin-based chemotherapy as neoadjuvant therapy.
We need to determine if any of the options match the known side effects of these drugs.
Let's search to confirm the side effects of cisplatin and other common bladder cancer chemotherapy drugs.
\texttt{</think>}

\texttt{<search>}
common side effects of cisplatin chemotherapy \texttt{</search>}

\texttt{<information>}
[Search Results regarding Cisplatin toxicity and mechanisms...]
\texttt{</information>}

\texttt{<think>}
From the information, cisplatin is known to cause ototoxicity, which aligns with the patient's symptoms of sensorineural hearing loss. Now let's check the mechanisms of action for the options:
Option (A): Inhibition of proteasome - This is associated with drugs like bortezomib and is not used for bladder cancer.
Option (B): Hyperstabilization of microtubules - This is the mechanism of taxanes (e.g., paclitaxel, docetaxel).
Option (C): Generation of free radicals - This is not specific to a single mechanism and is not commonly associated with ototoxicity.
Option (D): Cross-linking of DNA - This is the mechanism of alkylating agents (e.g., cyclophosphamide).
Cisplatin is an alkylating agent, and its mechanism involves cross-linking of DNA.
Therefore, the correct answer should be (D).
\texttt{</think>}

\texttt{<answer>}0\texttt{</answer>}

\end{tcolorbox}
\end{table} 
%%%%%%%%%%%%%%%%%%%%%%%%%%%%%%%%%%%%%%%%%%%%%%%%%%%%%%%%%%%%%%%%%%%%%%%%%%%%%%%
%%%%%%%%%%%%%%%%%%%%%%%%%%%%%%%%%%%%%%%%%%%%%%%%%%%%%%%%%%%%%%%%%%%%%%%%%%%%%%%

\end{document}